# Representation Requirements for Supporting Decision Model Formulation

**Tze-Yun Leong**
MIT Laboratory for Computer Science
545 Technology Square, room 420
Cambridge, MA 02139
(leong@lcs.mit.edu)

## Abstract

This paper outlines a methodology for analyzing the representational support for knowledge-based decision-modeling in a broad domain. A relevant set of inference patterns and knowledge types are identified. By comparing the analysis results to existing representations, some insights are gained into a design approach for integrating categorical and uncertain knowledge in a context-sensitive manner.

## 1 Introduction

Research in knowledge-based decision systems (KBDS) combines artificial intelligence and decision analysis techniques to solve problems involving choice and uncertainty. The dynamic decision-modeling approach in KBDS advocates that the decision models for different problems should be dynamically constructed from a knowledge base [Breese, 1989, Wellman, 1990a]. This approach facilitates scalability and reusability of the knowledge bases. Moreover, the resulting decision models are context-sensitive and include only the relevant information specific to the problems. To date, however, while much progress has been made in improving the algorithms for manipulating decision models, the automated model construction process remains to be formalized.

This paper characterizes the knowledge for supporting dynamic decision-modeling in medicine. Characterizing such knowledge illuminates the representational and computational requirements for automating decision analysis in a broad domain. Unlike previous efforts, instead of concentrating on the structural components of the decision *model* such as nodes, conditional probabilities, and influences, we focus on the ontological features of the decision *problem* such as contexts, classes of observed events, classes of available actions, classes of possible outcomes, temporal precedence, and probabilistic and contextual dependencies.

By gaining insights into the nature of decisions, this exercise serves as a step toward developing a formal methodology for requirement analysis and realizing a uniform representation framework for supporting dynamic decision-modeling in KBDS.

The following discussions are based on the general system architecture depicted in Figure 1. Given a problem description, the *planner* or *decision-maker* constructs a decision model by accessing information contained in the *knowledge base*. The domain and the decision-analytic components of the knowledge base are integrated by the *knowledge-base manager*, which also serves as an interface to the planner.

The decision models considered are *qualitative probabilistic networks* (QPNs) [Wellman, 1990b]. Since QPNs are the qualitative variants of influence diagrams, and since each influence diagram can be transformed into a decision tree, our results are expected to be generalizable to other decision models.

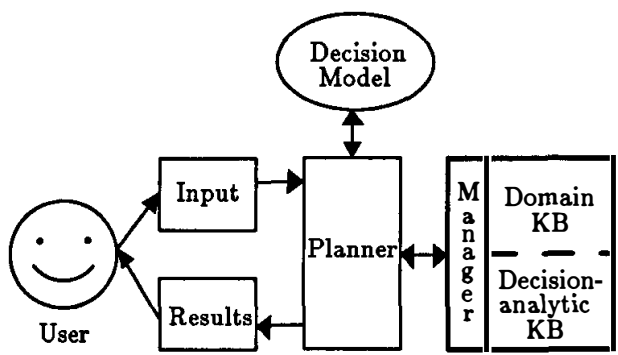

Figure 1: A Knowledge-Based Decision System

In the subsequent sections, we present a medical decision problem and examine the reasoning and representational issues involved in the decision analysis process. Some ideas on integrating context-sensitive categorical and uncertain knowledge will be explored and compared to relevant representation frameworks.

## 2  An Example

A simplified medical decision problem [Beck and Pauker, 1981, Tsevat *et al.*, 1989] is shown below:

*The patient is an 80 year-old woman. She complained of fainting and was found to have irregular heartbeats, or arrythmia. A diagnosis of cardiomyopathy, i.e., disorder of the heart muscles was made. Such a disorder usually leads to embolism, or formation of blood clots in the patient's body. The problem is to determine if anticoagulant therapy should be administered to reduce the chance of embolism, given the high risk of bleeding complications in the elderly.*

Each relevant event in the decision problem can be regarded as a *concept*, e.g., 80-year-old, cardiomyopathy, anticoagulant-therapy, etc. A concept is an *event* or a *random variable* in the probabilistic sense; it denotes an abstract description of an object, an attribute, a state of being or a process, depending on the circumstances.

## 3  The Decision Making Process

Given a set of input concepts, the goal for the proposed KBDS is to construct a decision model such as the one shown in Figure 2, and then evaluate the feasibility of the alternatives with respect to some criteria,e.g., life-expectancy, expected monetary value, etc.

More formally, the decision-analytic approach to decision making can be viewed as a five-step process: 1) Background characterization; 2) context establishment; 3) problem formulation; 4) model construction; and 5) Model evaluation.

### 3.1  Background Information Characterization

The process begins by classifying the input concepts into the variables concerned, the actions available, and the possible outcomes involved in a decision problem. In the clinical setting, the input concepts can usually be divided into six categories, as shown in Table 1 for our example.

The planner can characterize each input concept by asking questions like:

- Is fainting a kind of sign or symptom?
- Is cardiomyopathy a kind of disease?

To answer the above queries, the knowledge-base must support *categorizations* of the relevant domain concepts. A categorization is a grouping of concepts with similar descriptions in a particular dimension. Examples of such groupings include those induced by the specialization (AKO) relation, the decomposition (PARTOF) relation, etc.

Table 1: Characterized Background Information

| Category | Concepts |
|---|---|
| General history | 80 year old, female |
| Signs and Symptoms | Fainting, arrythmia |
| Laboratory findings | - |
| Diseases | Cardiomyopathy |
| Alternatives | Anticoagulant-therapy |
| Complications | Embolism, bleeding |

The characterized background information, however, is insufficient for formulating a decision model. For instance, in our example, the relationships among the input concepts are not explicitly stated, the two relevant kinds of embolism being considered, systemic and pulmonary embolisms, are not specified, and the evaluation criteria are not mentioned. The missing information, which may be related to the domain or the decision-analytic methodology, must be derived when necessary.

### 3.2  Domain Context Establishment

Establishing the context[1] means defining the task environment in which the problem is to be solved. This enables different problem situations to be considered and sets limits on the possible operations that can be applied to a given problem [Kassirer and Kopelman, 1987]. The context is selected with only a few clues [Kassirer and Gorry, 1978]. In the clinical setting, a context is usually indicated by a suspected disease, a syndrome, i.e., a set of signs and symptoms that convey special meanings,or a general diagnostic category, e.g., an acute respiratory disorder [Kassirer and Kopelman, 1987].

In our example, the clinical context is "cardiomyopathy in old-age." This context is established by identifying the suspected diseases and any conditions that might significantly affect their nature.

Given the characterized background information,identifying the suspected diseases simply involves looking them up in the set of input concepts. For now, we assume that other significant conditions, e.g., old-age in our example, are specified by an oracle. Recognizing these conditions automatically requires a very sophisticated planner, and the issues involved are outside the scope of this paper.

The main purpose of establishing a context is to al-low access to the context-sensitive information. For instance, in older patients, cardiomyopathy may have different manifestations and more severe complications than in younger patients, or in the presence of other diseases. Therefore, such context-sensitive knowledge must be expressible in the knowledge base.

---

[1]This is different from the *decision context* [Breese, 1989, Holtzman, 1989] which refers to all the assumptions, constraints, variables, and alternatives considered in the decision problem.

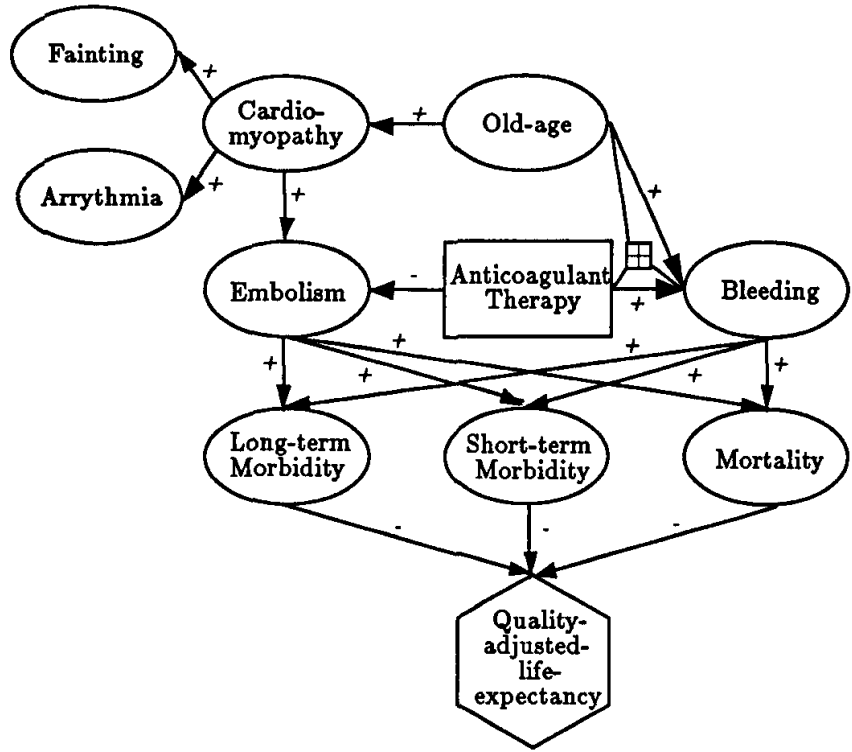

Figure 2: A QPN For The Example

### 3.3   Decision Problem Formulation

Guided by the characterized background information, a decision problem is formulated within the domain context by identifying:

- all or the most important diseases/hypotheses that may be involved;
- the relative significance of all these concepts;
- all or the most important possible outcomes/complications of these concepts;
- all or the most important actions available;
- the effects of the actions on the concepts and their outcomes and possible complications; and
- the evaluation criteria.

Table 2 shows all the relevant concepts in our example. "Pulmonary embolism" and "systemic embolism" are among the values of the corresponding "embolism" node in Figure 2.

Table 2: Concepts Involved in Decision Problem

| | |
|---|---|
| Old-age | Anticoagulant-therapy |
| Cardiomyopathy | Bleeding |
| Fainting | Long-term-morbidity |
| Arrythmia | Short-term-morbidity |
| Embolism | Mortality |
| Pulmonary-embolism | Quality-adjusted- |
| Systemic-embolism | life-expectancy |

These concepts are derived by asking questions like:

- What are the most common embolisms caused by cardiomyopathy?
- What are the other (if any) complications of anticoagulant-therapy?

To answer the above queries, the knowledge base must, in addition to supporting categorizations of the domain concepts, allow expression of the *interactions*, i.e., the correlational/influential/causal relations, among these concepts. The varying degrees of significance for all these relations in different con-texts must also be expressible in the knowledge base. This, together with the varying degrees of temporal and probabilistic dependencies among the interactions, would facilitate derivation of the most relevant information for the problem at hand.

## 3.4  Decision Model Construction

As mentioned, a decision model for our example is shown in Figure 2.

To construct such a decision model, its structure, e.g., nodes and links in a QPN, and its preference model, e.g., evaluation criteria such as morbidity, mortality, and monetary costs associated with utilities, must be inferrable from the knowledge base. The temporal constraints on the decision model structure, i.e., the order in which the concepts and their consequences are to be considered, should also be inferrable from the interactions of the underlying concepts. Hence, the construction involves asking questions like:

- How are the observable effects of the alternatives relate to the chance events?

- What are the outcomes that affect the evaluation criteria?

To support these queries, the relevant interactions among the concepts must be expressible in the knowledge base. These interactions involve both domain concepts and decision-analytic concepts, e.g., "presence of disease positively-influences morbidity."

## 3.5  Decision Model Evaluation

Upon completion, the decision model is evaluated by some procedure with respect to the evaluation criteria. Here, evaluation of a decision model refers to solving the model with procedures such as folding back of a decision tree, or graph reduction of a QPN. The evaluation criterion assumed in our example is quality-adjusted life expectancy, i.e., a measure of time remaining in a patient's life, taking into account the inconveniences caused by the illness (morbidity). Given a well-formed decision model, only procedural knowledge is needed in this step.

# 4  Summary of Inference Patterns and Representation Requirements

The above analysis shows that four types of general inference patterns are involved in the automated decision analysis process:

- (Q1) Does concept A related to concept B in $\mathcal{O}$

- (Q2) What are the concepts related to concept A in $\mathcal{O}$?

- (Q3) Does concept A relate to concept B by $i$?

- (Q4) What are the concepts related to concept A by $i$?

where $\mathcal{O}$ is a categorization and $i$ is an interaction.

Three types of knowledge are required to support these inferences: categorical knowledge, uncertain knowledge, and a notion of "context."

## 4.1  Categorical Knowledge

The categorical knowledge captures the definitional or structural relations of the concepts, allowing expression of facts such as: "cardiomyopathy is a kind of disease" and "pulmonary embolism is a kind of embolism." This type of knowledge should provide the system with the power of abstraction and inheritance. In other words, knowing a class of concepts would allow the planner to derive its subclasses, and vice versa. Furthermore, the generic description for a class of concepts can be specified at an appropriate level of abstraction; portions of this description can be inheritable by its subclasses or superclasses.

## 4.2  Uncertain Knowledge

The uncertain knowledge captures the interactions, i.e., the correlational, influential, or causal relations among the concepts, allowing expression of facts such as: "presence of anticoagulant-therapy negatively-influences presence of embolism" and "cardiomyopathy causes arrythmia." This type of knowledge should provide the system with the power of differentiation by accommodating a spectrum of temporal and probabilistic dependencies among the concepts. By comparing the relational strengths, the planner would be able to deduce the certainty and usefulness of the information derived from the knowledge base.

## 4.3  A Contextual Notion

In addition to the categorical and uncertain knowledge, a notion of "context" should be included in the knowledge base. This contextual notion has the following properties:

1. It sets a boundary on the relevant categorical and uncertain knowledge, and can be regarded as a focusing mechanism. This enables the planner to look for different information in different situations. For instance, the old-age of a patient would lead to the focus on a particular set of complications for cardiomyopathy and anticoagulant-therapy.

2. It allows differentiation of the relational significance, both categorical and interactional, among a set of concepts; the relative relevance and importance of the information can thus be distinguished in different situations. For example, bleeding is the most important complication of anticoagulant-therapy in the context of cardiomyopathy in old-age.

3. It is compositional and can be defined hierarchically. In other words, multiple, interacting contexts may coexist and a context can be defined within another context. For example, "cardiomyopathy" and "old-age" combine to form the context of "cardiomyopathy in old-age"; the latter, in turn, is a subcontext of "disease in old-age."

# 5    A Representation Design

We now propose a representation design that would meet the requirements for supporting the inferences Q1-Q4 in a context-sensitive manner.

## 5.1    Representation of Concepts

In our framework, a *concept* is an intensional description of the relational interpretation of an object, a state, a process, or an attribute of these phenomena. In other words, a concept reflects the salient features of the underlying phenomenon through a set of interactions with other concepts. These relevant concepts are called the *properties* of the concept being described. For example, the description of the concept **disease**[2] includes properties such as **severity**, **manifestation**, and **treatment**, as well as interactions such as **"presence-of-disease causes presence-of-manifestation-of-disease"** and **"presence-of-treatment-of-disease negatively-influences severity-of-disease."**

### 5.1.1    Properties of Concepts

The properties of a concept include its inherent qualities, characteristics, and other relevant concepts that constitute its description e.g., **size** (of a **tumor**) and **treatment** (of a **disease**). Each property is a concept[3] itself. Each property of a concept has a list of *values*, which are also concepts themselves.

To incorporate context-dependent information, a new concept can be derived from each property of a concept. For example, the concept **treatment-of-cardiomyopathy** is derived from the property **treatment** of **cardiomyopathy**. This new *derived-concept* has a description constrained by the concept it is derived from; the two concepts are related by the *context* (**CXT**) relation to be described below. Compositions of the **CXT** relation enable "chaining" of the derived-concepts, e.g., **duration-of-treatment-of-cardiomyopathy,**

**presence-of-complications-of-- treatment-of-cardiomyopathy**, etc. , are concepts formable in this way.

The properties of a concept in this framework are analogous to the *roles* in term subsumption languages and the *slots* in frame-based languages. The difference is that the properties alone do not completely describe a concept; they serve only as indices to the interactions that constitute the meaning of a concept. These interactions are expressed in terms of the corresponding derived-concepts, e.g., **"duration-of-treatment-of-cardiomyopathy negatively-influences severity-of-cardiomyopathy** is an interaction in the description of **cardiomyopathy.**

### 5.1.2    Interactions of Concepts

Each interaction between two concepts has two components: *temporal precedence*, with "known" or "unknown" as values, and *qualitative probabilistic influence* [Wellman, 1990b], with "positive," "negative," and "unknown" as values. The interactions can thus be expressed as four types of links in a network interpretation of our framework: *associational* links, which denote probabilistic correlation with an unknown type of influence and unknown temporal precedence; *precedence* links, which denote temporal precedence with unknown type of probabilistic influence; *influential* links, which denote conditional probabilistic dependencies; and *causal/inhibitive* links, which denote known temporal precedence in addition to known type of probabilistic influences.

## 5.2    Categorization of Concepts

The description of a concept can be constrained by a set of *categorizers*. A categorizer is a categorical or *class* relationship; it is a binary relation that specifies the properties and the interactions of a concept in terms of those of another concept. By imposing a partial order on the related concepts, a categorizer establishes a unique *perspective* for describing each concept. For example, a concept can be described as "a kind of" another concept or "a part of" another concept. Some common categorizers include the specialization (**AKO**) relation, the decomposition (**PARTOF**) relation, and the equivalence (**EQV**) relation.

All the concepts related by a categorizer are said to be in a *categorization*; some categorizations have hierarchical interpretations, while others are more naturally seen as networks. By knowing the position of a particular concept with respect to another concept in a categorization, the description of the former can be inferred from the latter. This descriptive inference in a categorization is called *inheritance*.

For instance, the specialization relation can be defined as follows:

---

[2]All concepts defined in our framework will be referenced in **typewriter type style.**

[3]Referred to as *property-concept* from now on.

**Definition .1 (Specialization)** *Let* $\mathcal{C}$ *be the set of all concepts. Let* $\Omega$ *be the set of categorizers. Let* $\mathcal{O}_\omega \subseteq \mathcal{C}$ *be the set of concepts in a categorization related by categorizer* $\omega \in \Omega$. *For all* $a, b \in \mathcal{C}$, *and for* $AKO \in \Omega$ *where* $AKO \subseteq \mathcal{C} \times \mathcal{C}$:

1. $AKO \stackrel{def}{=} \{(a,b) | a \subset b,$ *i.e.,* $\forall \alpha, \alpha \in a \implies \alpha \in b\}$.

2. *Let* $ako : \mathcal{C} \longrightarrow 2^{\mathcal{C}}$ *be a function defined on* $AKO$:

   $$ako(a) = \{b | (a, b) \in AKO\}.$$

Two major properties are observed for the **AKO** categorizer:

1. $a \in \mathcal{O}_{AKO} \iff \exists b, (a, b) \in AKO$ or $(b, a) \in AKO$.

2. The $AKO$ relation is irreflexive, asymmetric, and transitive.

3. The properties and interactions of the concepts are downward inheritable in the specialization hierarchy.

### 5.3 Context-Dependent Representation

The categorizers establish some general perspectives for describing a concept. For example, a **pulmonary-embolism** is a kind of **embolism** in general. The description of a concept in these general perspectives is further constrained by a set of *contexts*.

A context (**CXT**) relation can be thought of as a "meta-categorizer;" it is a binary relation that specifies the properties and the interactions of, and hence also the categorizers on a concept in accordance with those of another concept. For example, **treatment-of-cardiomyopathy** is specified as a kind of **treatment-of-disease** because **treatment-of-cardiomyopathy** is defined in the context of **cardiomyopathy**, and **cardiomyopathy** is a kind of **disease**. All concepts are described in some contexts; the descriptions that are valid in general are in the *universal* context. Therefore, the (**CXT**) relation facilitates representation of context-sensitive information, as mentioned earlier, by allowing chaining of derived-concepts and constraining their descriptions. The partial-ordering imposed by this relation forms a *context-hierarchy* of all the concepts in the knowledge base.

## 6   Supporting General Inferences

Based on the above representation framework, we shall now discuss how the knowledge base of the proposed KBDS would provide answers for the inferences Q1-Q4. In the following discussions:

1. Let $\mathcal{C}$ be the set of all concepts.

2. Let $\Omega = \{AKO, ...\}$ = the set of all categorizers.

3. Let $\mathcal{F}_\Omega = \{f_\omega | f_\omega$ is a function defined on $\omega, \forall \omega \in \Omega\} = \{ako, ...\}$ as defined in Section 5.2.

4. Let $\qquad \mathcal{I} \qquad = \qquad \{association,$ *precedence, positive-influence, negative-influence, cause, inhibitor*$\}$=the set of all interaction types.

5. $\forall i \in \mathcal{I}$, let $\mathcal{F}_\mathcal{I} = \{f_i | f_i$ is a function defined on $i\}$

6. $\forall f_i \in F_\mathcal{I}, i \in \mathcal{I}, a, b \in \mathcal{C}, f_i(a) = \{b | (a, b) \vee (b, a) \in i\}$.

- Q1: Does concept A relate to concept B by <categorizer>?

To find out if two concepts **A** and **B** are related in an categorization, let $\omega_0 \in \Omega$ be the categorizer in question.

$$\text{Answer}_{Q1} = \begin{cases} \text{yes} & \text{if } (A, B) \in \omega_0 \\ \text{no} & \text{otherwise.} \end{cases}$$

An example of the Q1 query is: Does **cardiomyopathy** related to **disease** by specialization? The answer is: yes.

- Q2: What are the concepts related to concept A by <categorizer>?

To find out the concepts related to a concept **A** in an categorization, again let $\omega_0 \in \Omega$ be the categorizer in question.

$$\text{Answer}_{Q2} = f_{\omega_0}(A).$$

An example of the Q1 query is: What are the concepts that are related to **embolism** by specialization? The answers are: **pumonary-embolism** and **systemic-embolism**.

- Q3: What are the concepts that relate to concept A by <interaction>?

To find out the concepts that directly interact with a concept **A** in an interaction, let $i_0 \in \mathcal{I}$ be the interaction in question.

$$\text{Answer}_{Q3} = f_{i_0}(A).$$

An example of the Q3 query is: What are the concepts that relate to **complication-of-anticoagulant--therapy** by positive-influence? The answer is: **presence-of-old-age**.

- Q4: Does concept A relate to concept B by <interaction>?

To find out whether two concepts **A** and tt **B** are involved in an interaction, again let $i_0 \in \mathcal{I}$ be the interaction in question.

$$\text{Answer}_{Q4} = \begin{cases} \text{yes} & \text{if } (A, B) \in i_0 \\ \text{no} & \text{otherwise.} \end{cases}$$

An example Q4 query is: Does **cardiomyopathy** relate to **fainting** by cause? (Read: Does **cardiomyopathy** cause **fainting**?) The answer is: yes.

For simplicity, all the answers to the above inferences assume a *closed world assumption*, i.e., a negative answer will be returned if a relation is not explicitly derivable from the knowledge base. The context-sensitivity of the answers, though not very obvious, is actually inherent from the underlying representation.

## 7    Related Work

The major shortcomings of the static decision-modeling approach, i.e., treating pre-enumerated decision models or templates as knowledge bases, result from the rigidity of the knowledge bases. Constrained by the structure of the decision models, e.g., nodes and links of a decision tree, such knowledge bases do not reflect the nature of the domain knowledge.

The different representations used in existing KBDS with the dynamic decision-modeling approach are not very satisfactory, either. The first order logic-like representations, such as those employed by Breese [1987, 1989], and Goldman and Charniak [1990], have no explicit hierarchical dimensions. In these frameworks, multi-level decision models are created by activation of a set of rules; limited contextual information is captured as conditional probabilities matrices in these rules.

In Wellman's [1990a] SUDO-PLANNER system, domain descriptions can be expressed in multiple levels of precision in this framework, thus facilitating decision-modeling in multiple levels of abstraction. The terminological component of this framework, however, is subjected to the limited expressiveness of most *term subsumption languages*. Moreover, the purely probabilistic nature of the effects or influences does not reflect the varying degrees of significance among the concepts with respect to the problem at hand. Although some contextual effects on the influences can be expressed in the *qualitative synergies* defined in QPN, there is no general mechanism for capturing contextual information in the whole framework.

Other relevant representation formalisms include those that incorporate an uncertainty model to a hierarchical representation framework. Most hierarchical representations are designed to support derivation of absolute or categorical answers. To support approximate reasoning, i.e., finding out facts that are not absolutely true or false, but *believed* to a certain degree, some efforts attempt to accommodate an uncertainty model by re-interpreting the semantics of a categorical representation, while others try to couple the two to form a coherent framework.

For instance, in the network representation developed by Lin and Goebel [1990], both subsumption and causal relationships are expressible. Probabilistic interpretations are given to parts of the causal network, called the *scenarios*. These scenarios can be considered as contexts with different probability distributions. Although the scenarios are not hierarchically arranged, their probabilistic rankings are preserved across the subsumption relationships. Nevertheless, this network formalism does not allow the properties, and hence the nature of each node or concept to be explicitly represented.

Yen and Bonissone's [1990] work attempts to generalize the semantics of term subsumption languages with an approximate reasoning model, such as fuzzy logic or possibility theory, to support plausible inferences. Non-definitional relations among the concepts, however, are not expressible in these frameworks. There is also no general notion of context-dependent definitions.

Saffiotti's [1990] hybrid framework, on the other hand, integrates a component that deals with absolute or categorical knowledge and another with the uncertainty of this knowledge. Any formal representation formalism and uncertainty model may constitute the two components in the framework, e.g., first-order logic with Dempster-Shafer theory, term subsumption language with probability theory, etc. We believe this work is an important step toward the theoretical foundations of integrating categorical and uncertain knowledge. The expressiveness and hence the usefulness of the framework, however, depend solely on the component formalisms.

## 8    Discussion and Conclusion

To support dynamic decision-modeling, the structure of the knowledge base must reflect the nature of both the decision problem and the domain knowledge. In particular, the underlying representation must neither be restricted by the structural components of the decision models, e.g., nodes and links of an influence diagram, nor their evaluation mechanisms, e.g., folding back of a decision tree. By focusing on the ontology of a decision problem, we have identified a set of inference patterns and knowledge types for supporting automated construction of decision models in medicine.

The brief survey on existing representations has shed some light on a design approach for integrating categorical and uncertain knowledge in a context-sensitive manner. We believe such an integration calls for a framework with a terminological component, an assertional component, and a network interpretation. By capturing the context notion via partitioning the network, this framework would allow us to establish tax-onomies of structured concepts, state the facts, i.e., the interactions among the concepts, and answer questions about these relations.

We have sketched a design outline of such a representation in this paper; a more detailed exposition is described elsewhere [Leong, 1991]. Many important issues, however, are yet to be explored. In particular, the notion of "context" needs to be more formally defined, many interesting problems arise in the context-sensitive inheritance patterns of the categorical relations, and the context-sensitive probabilistic semantics of the interactions needs to be generalized. Careful examination of these issues, we believe, will lead to the formalization of both the automated decision model formulation process and the domain and decision-analytic knowledge involved.

### Acknowledgments

The author would like to thank Peter Szolovits for advice on the project, Mike Wellman for many helpful discussions, Jon Doyle for comments on the mathematical definitions and LaTeX formatting, and the anonymous referees for suggestions on the presentation of this paper.

This research was supported by the National Institutes of Health grant no. 5 R01 LM04493 from the National Library of Medicine.